\documentclass[journal]{IEEEtran}
\usepackage{cite}

\ifCLASSINFOpdf
  \usepackage[pdftex]{graphicx}
\else
\fi

\usepackage{amsmath}
\usepackage{siunitx}
\begin{document}
%
\title{Comfort-oriented driving: performance comparison between human drivers and motion planners}
%
%
%

\author{Yanggu~Zheng$^{1}$,
        Barys~Shyrokau$^{1}$,
        and~Tamas~Keviczky$^{2}$
\thanks{The research leading to these results has received funding from the European Union Horizon 2020 Framework Program, Marie Sklodowska-Curie actions, under grant agreement no. 872907.}
\thanks{$^{1}$Yanggu Zheng and Barys Shyrokau are with the Department of Cognitive Robotics, Faculty of Mechanical, Maritime and Materials Engineering,
        Delft University of Technology, 2628CD, The Netherlands
        {\tt\small y.zheng-2@tudelft.nl, b.shyrokau@tudelft.nl}.}
\thanks{$^{2}$Tamas Keviczky is with Delft Center for Systems and Control, Faculty of Mechanical, Maritime and Materials Engineering,
        Delft University of Technology, 2628CD, The Netherlands
        {\tt\small t.keviczky@tudelft.nl}}}

%
%

\markboth{arXiv preprint, submitted to IEEE T-ITS in Jan. 2023}%
{Zheng \MakeLowercase{\textit{et al.}}: }
%



\maketitle

\begin{abstract}
Motion planning is a fundamental component in automated vehicles. It influences the comfort and time efficiency of the ride. Despite a vast collection of studies working towards improving motion comfort in self-driving cars, little attention has been paid to the performance of human drivers as a baseline. In this paper, we present an experimental study conducted on a public road using an instrumented vehicle to investigate how human drivers balance comfort and time efficiency. The human driving data is compared with two optimization-based motion planners that we developed in the past. In situations when there is no difference in travel times, human drivers incurred an average of 23.5\% more energy in the longitudinal and lateral acceleration signals than the motion planner that minimizes accelerations. In terms of frequency-weighted acceleration energy, an indicator correlated with the incidence of motion sickness, the average performance deficiency rises to 70.2\%. Frequency-domain analysis reveals that human drivers exhibit more longitudinal oscillations in the frequency range of 0.2-1 Hz and more lateral oscillations in the frequency range of up to 0.2 Hz. This is reflected in time-domain data features such as less smooth speed profiles and higher velocities for long turns. The performance difference also partly results from several practical matters and additional factors considered by human drivers when planning and controlling vehicle motion. The driving data collected in this study provides a performance baseline for motion planning algorithms to compare with and can be further exploited to deepen the understanding of human drivers.
\end{abstract}

\begin{IEEEkeywords}
motion planning, motion comfort, motion sickness, human drivers, naturalistic driving
\end{IEEEkeywords}

%
\IEEEpeerreviewmaketitle

\section{Introduction}\label{sec:intro}
\IEEEPARstart{A}{dvantages} of automated vehicles (AVs) are highly attractive. Society could benefit from this emerging technology in a wide spectrum of aspects including but not limited to improving safety, providing mobility to individuals with special needs, and reducing energy consumption and traffic congestion. With automation handling the driving task independently, the occupants could utilize the travel time for other activities, which implies considerable societal benefit in terms of productivity. However, AVs must first demonstrate their advantages convincingly in order to be accepted by the public. It is argued that only matching the performance of human drivers is insufficient for promoting AVs, and a significant improvement is necessary instead\cite{liu2020self}. In terms of safety, AVs can potentially outperform human drivers with their advanced sensing capability. Human drivers rely on their eyes to obtain information from the visible light spectrum with an estimation of depth, which is comparable to the stereo-vision camera used on most AVs. Meanwhile, AVs could be equipped with a variety of additional sensors including LiDAR, radar, and thermal cameras to simultaneously exploit different sources of information. This promotes better detection performance in adverse visibility conditions, e.g. in darkness, fog, or heavy precipitation. Some suggested that AVs could drive safely at a higher speed in such circumstances \cite{schoettle2017sensor}. Other works have exploited the reflection of radar beams to achieve early detection of crossing pedestrians before they become visible, which provides more time for AVs to react safely \cite{palffy2022detecting}. Besides, AVs are expected to have more accurate control of vehicle motion, especially in safety-critical situations. Most of the current active safety systems assist human drivers and have already contributed to a significant reduction in crash rates \cite{lie2004effectiveness,lyckegaard2015effectiveness}. Given full control over vehicle motion, AVs are believed to further enhance road safety by preventing accidents due to loss of stability. Furthermore, AVs could help in scenarios involving interaction between multiple vehicles such as driving in a platoon \cite{wang2019benefits}, merging from a ramp \cite{gao2021optimal}, or crossing unsignalized intersections \cite{kamal2014vehicle}. These applications could potentially reduce fuel consumption \cite{hussein2021vehicle} and improve traffic efficiency \cite{vcivcic2021coordinating}. \par

Comfort is another important aspect that AVs are expected to improve over human-driven vehicles. It is the foundation of the societal benefits derived from the travel time saved by not having to perform the driving task. Without being actively engaged with vehicle control, the occupant who used to be the driver becomes equally susceptible to motion sickness as the passengers \cite{wada2016motion}. The susceptibility increases further when they perform secondary activities that require visual attention because the lack of external visual cues hampers their ability to anticipate vehicle motion \cite{irmak2021objective}. When motion sickness symptoms arise, the occupants exhibit a drop in performance and willingness in cognitive tasks \cite{smyth2018motion}. This in return undermines part of the productivity boost by AVs. The paradox can be solved only if AVs can achieve a high level of motion comfort. In a survey study, several approaches to motion planning were labeled as good for comfort, most of which fall in the category of interpolated curves \cite{gonzalez2015review}. Some other relevant studies aim at ensuring continuous curvature or curvature rate, which could be achieved by constructing vehicle trajectories using clothoid segments \cite{silva2018clothoid}. State lattices and motion primitives have also been explored in order to ensure smooth motion within a sampling- or optimization-based planning framework \cite{bottasso2008path, mcnaughton2011motion, mischinger2018towards}. A majority of these studies only demonstrated the benefit qualitatively based on the smoothness. We believe it is less meaningful to discuss the motion comfort of a vehicle without considering the factor of time. Typically, the movement of the vehicle exerts inertial accelerations on the passengers when the speed or direction changes. The inertial accelerations are considered the main source of discomfort as they are perceived as disturbances to the passengers' body posture and can cause motion sickness \cite{turner1999motionA}. The goal of reducing acceleration can be achieved by driving at a minimal speed but this leads to longer travel time and can potentially cause danger to other road users. Factoring in time efficiency in comfort-oriented motion planning could prevent such undesirable behaviors and make AVs socially acceptable. The relative importance between comfort and time efficiency has been studied on multiple occasions as a critical design parameter \cite{shin2018kinodynamic, zheng20223dop}. Meanwhile, the topic of mitigating motion sickness in motion planning has been addressed which also included time as part of the objective function in the formulation of the optimization problem \cite{htike2021fundamentals}. \par

It is a convincing way to validate the benefits of AV by directly comparing them with human drivers as a baseline. The most convenient approach to this is to employ an artificial driver model. However, most driver models only focus on a specific domain of application such as steering \cite{qu2014switching} or traffic following \cite{lindorfer2018modeling}. To our knowledge, no driver model is generally representative of human drivers. Furthermore, the development of such models still relies on experimental research on real human drivers. Moving-base driving simulators have been widely used in order to achieve this, especially when investigating safety-critical scenarios \cite{li2019drivers, quante2021human} because it minimizes the risk of causing physical harm to the participants and test equipment. However, driving simulators also have their limitations. A positive correlation has been found between better performance on a driving simulator and a higher pass rate on the driving test \cite{de2009relationships} but it does not imply a matching between the magnitudes of the vehicle motion in the virtual and real world. The motion cueing of the simulator base is limited by the actuation constraints. Hence the vehicle motion perceived by a test driver is not as aggressive as in the real world despite the efforts in improving the motion cueing algorithms \cite{khusro2020mpc}. Besides, the visualization of the virtual world could be of insufficient fidelity. Even with realistic graphics, the loss of depth perception could still cause the driver to perform differently. Moreover, it is not guaranteed that a higher-fidelity simulator leads to more valid experiment results reflective of real-world driving performance. There are further issues with the need for learning and adapting to a driving simulator before the participant could feel comfortable and confident to drive it. The importance of adaptation has been shown with experimental studies \cite{sahami2013drivers, ronen2013adaptation}. When recruiting from the general public, the majority of participants might have very little to no experience with a driving simulator. If given not enough time and instructions, there is a risk that the results are not representative of the drivers' actual performance. In order to gather driving data where drivers behave in a naturalistic fashion, some researchers recorded vehicle trajectories using parked vehicles with LiDAR mounted on the roof \cite{zyner2019acfr} or drones with an onboard camera \cite{krajewski2020round, breuer2020opendd}. These methods are convenient when recording at a specific location but cannot provide accurate measurements of vehicle accelerations \cite{zheng2021comfort}. Therefore, an experiment with participants driving a real vehicle is the most suitable option for our study. This approach involves significantly more preparation and data-processing effort and needs to accept a lack of control over the environment but the benefits still outweigh the disadvantages. Compared with simulator-based experiments, it allows the participants to drive in a more natural setup so that the results are not influenced by the factors such as the fidelity of the simulator itself. Compared with remote sensing, it offers the freedom to choose sensors that suits the type of motion data that is the most relevant. It further allows us to communicate the experiment's goal so that the participants are driving attentively and deliberately. This helps in finding an upper limit of human driving performance. A comparison to this performance level of human drivers could evidently support the claim of the advantage of AVs. \par

The contributions of our study include:
\begin{itemize}
    \item A real-world experiment to collect data for measuring human driving performance in terms of motion comfort and time efficiency.
    \item A comparison study between human drivers and optimization-based motion planners aiming at improving comfort or mitigating motion sickness while maintaining time efficiency.
    \item An analysis of the frequency components in human driving patterns in comparison with the optimization-based motion planners.
\end{itemize}


The rest of this paper is organized as follows. Section II explains the design of the experiment, including the choice of vehicle and route, recruitment of participants, and processing of data. Section III introduces the two optimization-based motion planners that we will use to compare with the human drivers. The actual comparison between human drivers and motion planners is presented in Section IV. Finally, the contributions and limitations of this work are summarized in Section V, where we also make recommendations for future research. \par
\section{On-road Experiment}\label{sec:exp}

\begin{figure*}
    \centering
    \includegraphics[trim={0pt 40pt 40pt 50pt}, clip, width=\linewidth]{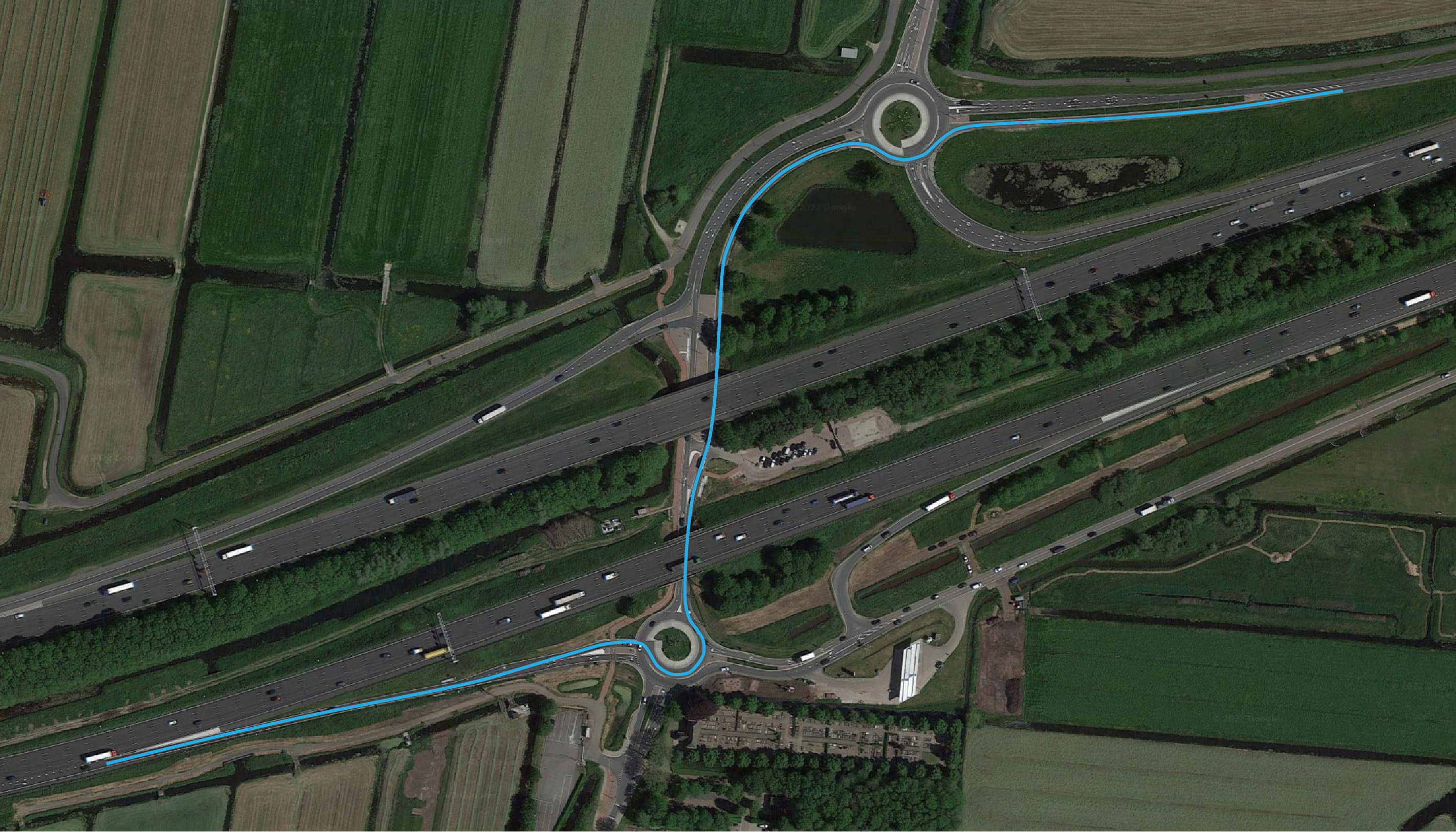}
    \caption{Illustration of the test route on the satellite image.}
    \label{fig:route}
\end{figure*}

\subsection{Procedures}\label{exp-procedures}

We recruited volunteers publicly to participate in the experiment. The participants were required to drive an instrumented vehicle through a part of the public road as the test route. They were specifically asked to drive smoothly and fast while staying within the range of aggressiveness they are confident with. They were briefed about the experiment and non-sensitive personal data including gender, age range, driving frequency and experience, and familiarity with the route were collected. This is to verify if there is a statistical bias in the group. After getting seated in the vehicle, the participants were given sufficient time to familiarize themselves with the driver interface before they started driving. While driving to the starting point of the test route, they were given the opportunity to get accustomed to the vehicle's handling, e.g. the steering and pedal feel and the vehicle's response. During the actual test, the participants drove the vehicle through a pre-defined route while the position and acceleration data were being recorded. It is obvious that interacting with other road users has a negative impact on the measured driving performance characterized by the duration of the drive and the total accelerations. Therefore, we only conducted the experiment outside rush hours. All tests were in one of the following time slots: 9:45-12:00, 13:15-15:30, or 18:15-20:30. Furthermore, each participant needed to drive the route twice in order to increase the chance of capturing a run without being influenced by other vehicles in the traffic. \par

\subsection{Test Route}\label{exp-route}

A challenging scenario is needed to sufficiently expose the variance in driving performance. Ideally, the test route requires the vehicle to change its speed across a wide range and negotiate a fair amount of turns. In addition, the distance and duration should be reasonably short and with a minimal trivial portion where most drivers would behave very similarly by driving almost straight and sticking to the speed limit. For these reasons, we chose the test route located to the west of Woerden, the Netherlands. As depicted in Fig. \ref{fig:route}, the start of the route is on the exit ramp of motorway A12 on the side of eastbound traffic. At the end of the ramp is a double-lane roundabout (further referred to as RB1) where the vehicle needs to turn to the left by taking the 3rd exit. Following up is an intermediate sector where the vehicle follows the road and passes 3 turns in a right-left-right order (further referred to as RLR). The vehicle then enters the second roundabout (further referred to as RB2) where it takes the exit towards distributor road N420 further to the east. After departing the roundabout, the vehicle is allowed to accelerate until the speed limit is reached. Through the test route, there are two occasions where the test vehicle may need to yield to other road users, namely when the vehicle enters RB1 and RB2. For the rest of the route, the test vehicle travels on a priority lane where it would not be impeded by other road users according to traffic rules. \par

\subsection{Participants}\label{exp-participants}

The majority of participants were recruited from the neighboring area using flyers distributed to residential addresses and companies. This potentially promotes a more inclusive group than recruiting only on campus. Besides, the test route is an important part of the daily commute of the local population. Most participants were familiar with the route prior to the experiment and therefore did not require much learning and practice in order to perform to the best of their capability. A total of 16 participants registered for the experiment. According to the collected non-sensitive personal data, only 2 participants reported that they were not familiar with the test route and 3 reported that they drive a car less frequently than once a week. Meanwhile, 12 participants have been in possession of a driving license for over 10 years and three reported over 4 years. In terms of age, 5 participants were under 30 at the time of the experiment, 2 were above 60, and the rest were between 31 and 59. The largest bias was present in terms of gender with only 4 female participants in the group. Given these figures, we are confident that the group is representative of a skillful human driver with average experience. \par

\subsection{Vehicle and Data Acquisition}\label{exp-vehicle}

\begin{figure}
    \centering
    \includegraphics[width=\linewidth]{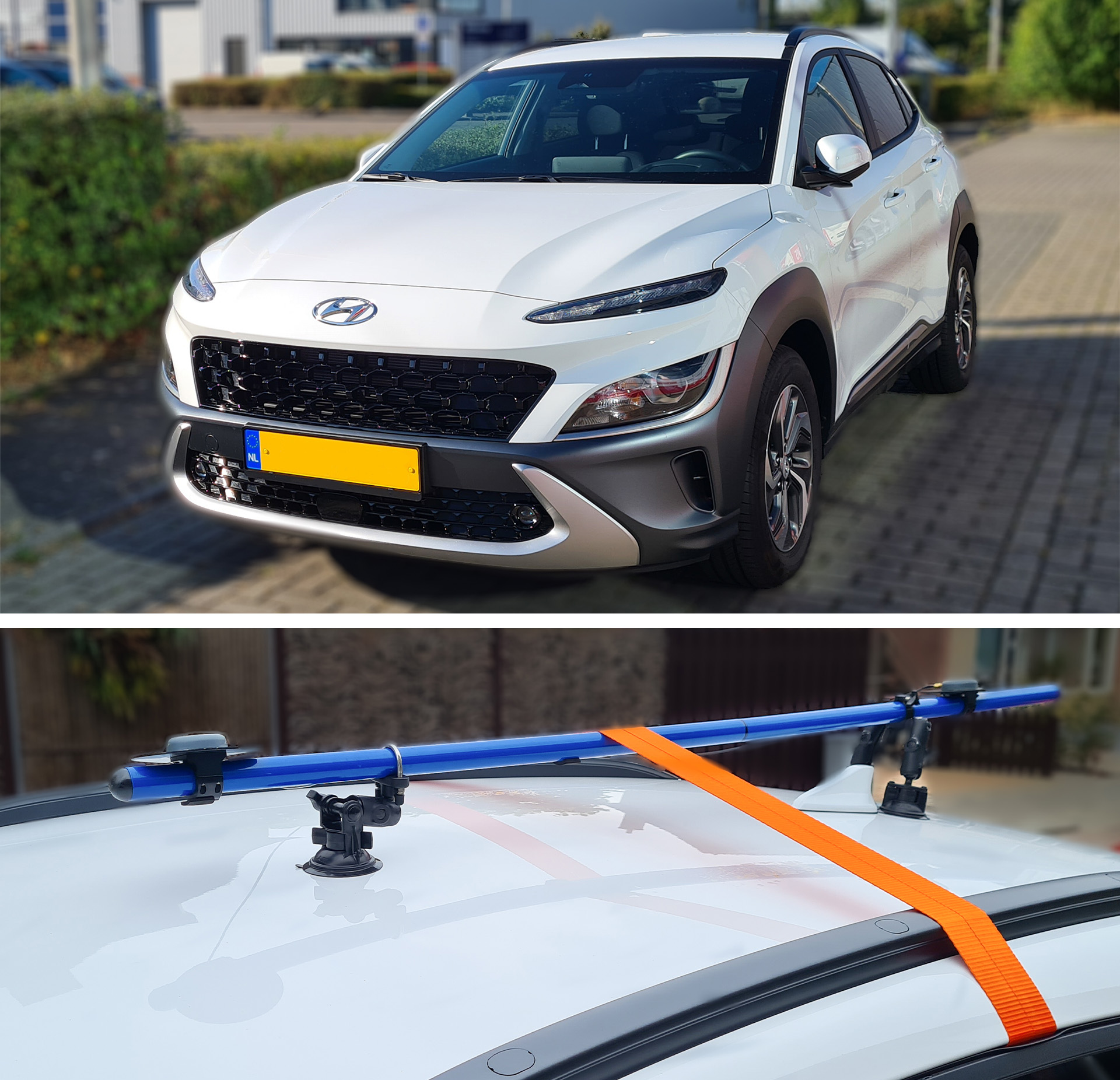}
    \caption{Instrumented vehicle used in the experiment}
    \label{fig:vehicle}
\end{figure}

The experiments were conducted on a Hyundai Kona that features an automatic transmission (AT) and a hybrid powertrain. The AT was chosen on purpose in order to reduce the participants' workload during the experiment. It eliminates the need to consider gear changing and to learn the clutch pedal characteristics. The hybrid powertrain was a compromised option as no combustion engine-powered vehicle with AT was available to us. To minimize the impact of driving with a less familiar powertrain, the regenerative braking is set to the minimum so that releasing the throttle pedal does not cause a strong deceleration. \par
For data acquisition, we used the Racelogic VBOX GPS speed sensor and the XSENS inertial measurement unit (IMU). The speed sensor had two roof-mounted GPS antennas secured with vacuum cups and tension straps. The antennas were aligned on a horizontal plane and with the centerline of the vehicle. According to \cite{heydinger1999measured}, a passenger car typically has its center of gravity located behind the front axle by 35-50\% of the wheelbase, and at 35-40\% of the overall vehicle height. Therefore, we mounted the IMU onto the bottom of the central glove box, which is the closest horizontal surface to the center of gravity according to the aforementioned figures. \par
With the GPS sensor, the recording of each test run was triggered automatically when the vehicle crossed a predefined gate defined with GPS coordinates. This ensured the reliable timing of the test run. However, the IMU sensor could not share this trigger signal and therefore had to be operated manually based on some landmark features. In practice, we observed a misalignment of less than \SI{0.2}{\second} when comparing the recorded data, which was compensated for when processing the data. The surrounding environment of the test route posed further challenges to the GPS sensor. Because the experiment was conducted on open public roads, the GPS signal was occasionally blocked by other vehicles of a larger size (e.g., semi-trailer trucks) as well as static objects such as trees and lamp posts. Moreover, the test route included an underpass across the motorway A4 where the connection to satellites was lost for approximately \SI{15}{\second}. The vehicle trajectory for this period of time could only be estimated with IMU measurements. \par

\subsection{Data Processing}\label{exp-data}
We reconstructed the motion of the vehicle using both sources of information, i.e., from GPS and IMU. As mentioned above, there were certain practical challenges in estimating the actual vehicle trajectory. An optimization-based processing method was developed in order to overcome these challenges. In principle, our method finds a motion profile by minimizing a cost function that includes the error between the measured and estimated positions and accelerations: 
\begin{equation}
    J({X_\text{est}}) = {w_1}{J_\text{GPS}} + {w_2}{J_\text{IMU}}
\end{equation}
Where,
\begin{equation}
    \begin{array}{l}
    {J_\text{GPS}} = \sum\limits_{k = 1}^N {\left( {{{\left( {{x_{\text{GPS},k}} - x_k} \right)}^2} + {{\left( {{y_{\text{GPS},k}} - y_k} \right)}^2}} \right)} \\
    {J_\text{IMU}} = \sum\limits_{k = 1}^N {\left( {{{\left( {{a_{x,\text{IMU},k}} - {a_{x,k}}} \right)}^2} + {{\left( {{a_{y,\text{IMU},k}} - {a_{y,k}}} \right)}^2}} \right)}
    \end{array}
\end{equation}
The decision variable $X_\text{est}$ in the optimization problem includes heading angles and velocities along with the initial position: 
\begin{equation}
    {X_{est}} = \left[ {x_0},{y_0},{{\psi _1} \cdots {\psi _N},{v_1} \cdots {v_N}} \right]
\end{equation}
The positions and accelerations can be calculated based on the decision variables as follows: 
\begin{equation}
    \begin{array}{rcl}
    {x_{k + 1}} & = & {x_k} + {v_k}{T_s}\cos {\psi _k}\\
    {y_{k + 1}} & = & {y_k} + {v_k}{T_s}\sin {\psi _k}\\
    {a_{x,k}} & = & \left( {{v_{k + 1}} - {v_k}} \right){f_s}\\
    {a_{y,k}} & = & {v_k}\left( {{\psi _{k + 1}} - {\psi _k}} \right){f_s}
\end{array}
\end{equation}

\noindent As mentioned in Section \ref{exp-vehicle}, the obstruction of the GPS signal causes the positional measurements to be less reliable. Hence the IMU measurement should be trusted more by placing a larger weight on the acceleration error. For the most part of the maneuver, we chose a weight distribution of $w_1=1,w_2=5$, which accounts for the relative scale between position and acceleration errors. For the period where the connection is lost, however, zero weight is given to the position error while the acceleration error is penalized alone with $w_2=10$. This does not mean that the estimated accelerations during this section of the route will match the IMU measurements exactly. The solution to the optimization problem still needs to balance between the two error types for the other parts of the test run.\par

\section{Optimization-based Motion Planners}\label{sec:planner}
\subsection{Problem Formulation}
We provide two motion planner variants from our past research for the purpose of comparison \cite{zheng2023mitigating}. They share the same structure and formulation while the difference lies in the objective function of the underlying optimization problem:

\begin{equation}
    \begin{array}{rc}
        \text{min:} & J_\text{Planner}(\textbf{X}) \\
        \text{where:} & \textbf{X} = \left[ {{y_1} \ldots {y_N},{v_1} \ldots {v_N}} \right] \\
        \text{s.t.:} & {y_{\min }} \le {{y_1} \ldots {y_N}} \le {y_{\max }}\\
        & {v_{\min }} \le {{v_1} \ldots {v_N}} \le {v_{\max }}\\
    \end{array}
\end{equation}

\noindent where the decision variables include $y_{k}$, the relative position of the vehicle with respect to the lane center, and $v_k$, the driving speed. The motion is defined by $N$ waypoints, hence the number of decision variables is $2N$. \par

The motion plan is defined under the assumption that the vehicle travels in a permissible motion corridor on well-paved road surfaces. The vehicle trajectory is constructed by connecting a string of waypoints. Each waypoint is located relative to a corresponding station distributed along the center of the driving lane. The relative position of the waypoint is constrained on the local lateral axis across the normal driving direction. A velocity is further assigned to each waypoint so that the vehicle motion is defined both in space and in time. This enables the calculation of accelerations, which is fundamental for evaluating the objective function. The detailed steps of calculation can be found in \cite{zheng20223dop}. \par
\subsection{Objectives}
The objective function is designed to achieve a balance between motion comfort and time efficiency. It concerns a measure of comfort plus a weighted total travel time:

\begin{equation}
    J_\text{Planner} = J_\text{Comfort} + WT
\end{equation}

\noindent Including time in the objective function effectively prevents the motion planner from commanding a minimal permissible velocity and the weighting term helps customize the comfort level. The comfort is measured by the total energy of the acceleration signal. As mentioned above, we provide here two variants of the objective function. In one variant, the acceleration signal is penalized as-is:

\begin{equation}
    {J_\text{Comfort, MA}} = \sum\limits_{k = 1}^N {\left( {a_{x,k}^2 + a_{y,k}^2} \right)\Delta {t_k}}
\end{equation}

\noindent This reflects the goal of improving general motion comfort because the accelerations are effectively the disturbance exerted on the passengers. The subscript MA is short for minimal acceleration. In the other variant, we penalize the frequency-weighted accelerations in the interest of mitigating motion sickness:

\begin{equation}
    {J_\text{Comfort, MS1}} = \sum\limits_{k = 1}^N {\left( {a_{xf,k}^2 + a_{yf,k}^2} \right)\Delta {t_k}}
\end{equation}

\noindent Here, the subscript MS is abbreviated from motion sickness. The frequency weighting is achieved by applying a band-pass filter to the longitudinal and lateral accelerations using the recommended values from the literature for the cut-off frequencies \cite{donohew2004motion, golding2001motion}. The filter is first constructed in the form of a continuous-time transfer function:
\begin{equation}
    H\left( s \right) = \frac{{{a_{\text{fil}}}\left( s \right)}}{{{a_{\text{act}}}\left( s \right)}} = \frac{1}{{{\tau _1}s + 1}}\frac{s}{{{\tau _2}s + 1}}
\end{equation}
Where $\tau_1$ and $\tau_2$ are the time constants corresponding to the desired cut-off frequencies. For zero initial conditions, the equivalent continuous-time state-space matrices of the filter are: 
\begin{equation}
\begin{array}{l}
A = \left( {\begin{array}{*{20}{c}}
{ - {\tau _1}^{ - 1} - {\tau _1}^{ - 1}}&1\\
{ - {\tau _1}^{ - 1}{\tau _2}^{ - 1}}&0
\end{array}} \right)\\
B = \left( {\begin{array}{*{20}{c}}
{{\tau _1}^{ - 1}{\tau _2}^{ - 1}}\\
0
\end{array}} \right)\\
C = \left( {\begin{array}{*{20}{c}}
1&0
\end{array}} \right)
\end{array}
\end{equation}
Given a time step of $\Delta t$ and assuming zero-order hold for the input, the filter can be converted to a discrete-time state-space model:
\begin{equation}
\begin{array}{l}
{x_{k + 1}} = {A_d}{x_k} + {B_d}{a_{\text{act},k}}\\
{a_{\text{fil},k}} = {C_d}{x_k}
\end{array}
\end{equation}
\noindent Where,
\begin{equation}
\begin{array}{l}
{A_d} = {e^{A\Delta t}}\\
{B_d} = {A^{ - 1}}\left( {{A_d} - I} \right)B\\
{C_d} = C
\end{array}
\end{equation}
The filter is initialized with zeros for the states at the beginning of the motion. At every step of movement, an acceleration input $a_{\text{act},k}$ and a step time $\Delta t$ is calculated. Consequently, it is possible to determine the matrices $A_d$, $B_d$, and $C_d$ accordingly. Especially, $A_d$ can be evaluated by first diagonalizing $A$ as:
\begin{equation}
    \Omega = {P^{ - 1}}AP
\end{equation}
\noindent For a diagonalizable matrix, its exponential is equal to:
\begin{equation}
    {e^{A\Delta t}} = P{e^{\Omega \Delta t}}{P^{ - 1}}
\end{equation}
\noindent Then for the diagonal matrix $\Omega \Delta t$, its exponential is calculated as:
\begin{equation}
    {e^{\Omega \Delta t}} = \left( {\begin{array}{*{20}{c}}
    {{e^{{\omega_{11}}\Delta t}}}&0\\
    0&{{e^{\omega_{22}\Delta t}}}
    \end{array}} \right)
\end{equation}
Using these steps, the frequency-weighted acceleration can be computed given a time-stamped series of actual acceleration. At the end of the series, however, the internal states of the filter do not immediately return to zero due to the slow dynamics. Therefore, the output of the filter continues to be penalized for a cooldown period of \SI{30}{\second} with zero input:

\begin{equation}
    {J_\text{Comfort, MS2}} = \sum\limits_{k = N+1}^{N+N_\text{Tail}} {\left( {a_{xf,k}^2 + a_{yf,k}^2} \right)\Delta {t_k}}
\end{equation}

\noindent Without such a penalty, the motion planner could command a more aggressive turning and change of speed at the end of the maneuver, which in reality should be avoided. The total comfort term in the sickness-mitigating variant of the motion planner is:

\begin{equation}
    {J_\text{Comfort, MS}} ={J_\text{Comfort, MS1}} + {J_\text{Comfort, MS2}}
\end{equation}

The motion planner variant that minimizes a cost function containing $J_\text{Comfort, MA}$ will be referred to as the MA planner, and $J_\text{Comfort, MS}$ as the MS planner. The optimized motion plans from these variants will be used for comparison purposes with human driving data. \par

\noindent 
\section{Results}\label{sec:result}

\subsection{Reconstruction of test runs}
\begin{figure}
    \centering
    \includegraphics[trim={20pt 0pt 20pt 10pt}, clip, width=\linewidth]{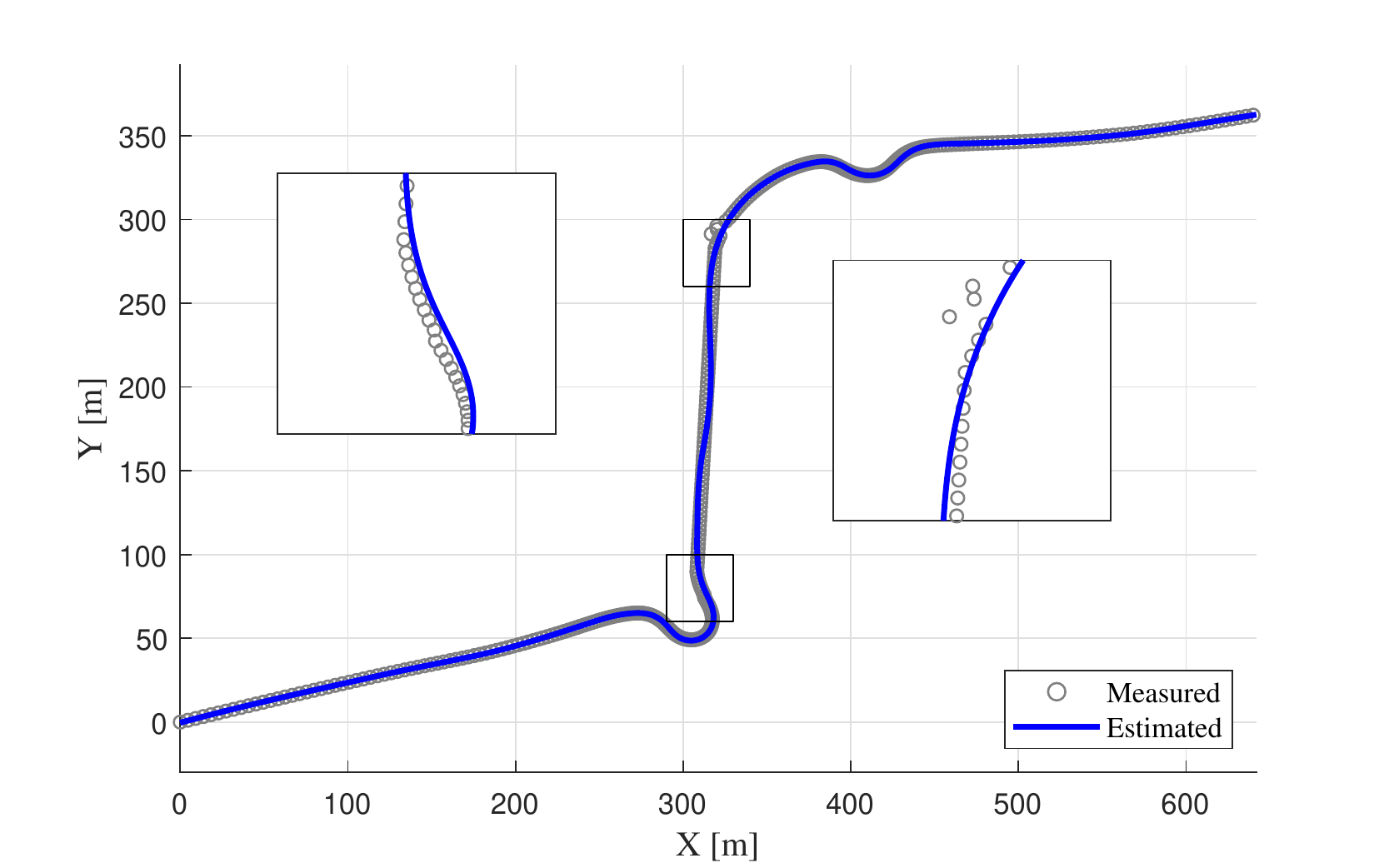}
    \caption{An example of the motion reconstruction method: estimated path vs. GPS measurement.}
    \label{fig:traj_est}
\end{figure}

\begin{figure}
    \centering
    \includegraphics[trim={20pt 20pt 20pt 30pt}, clip, width=\linewidth]{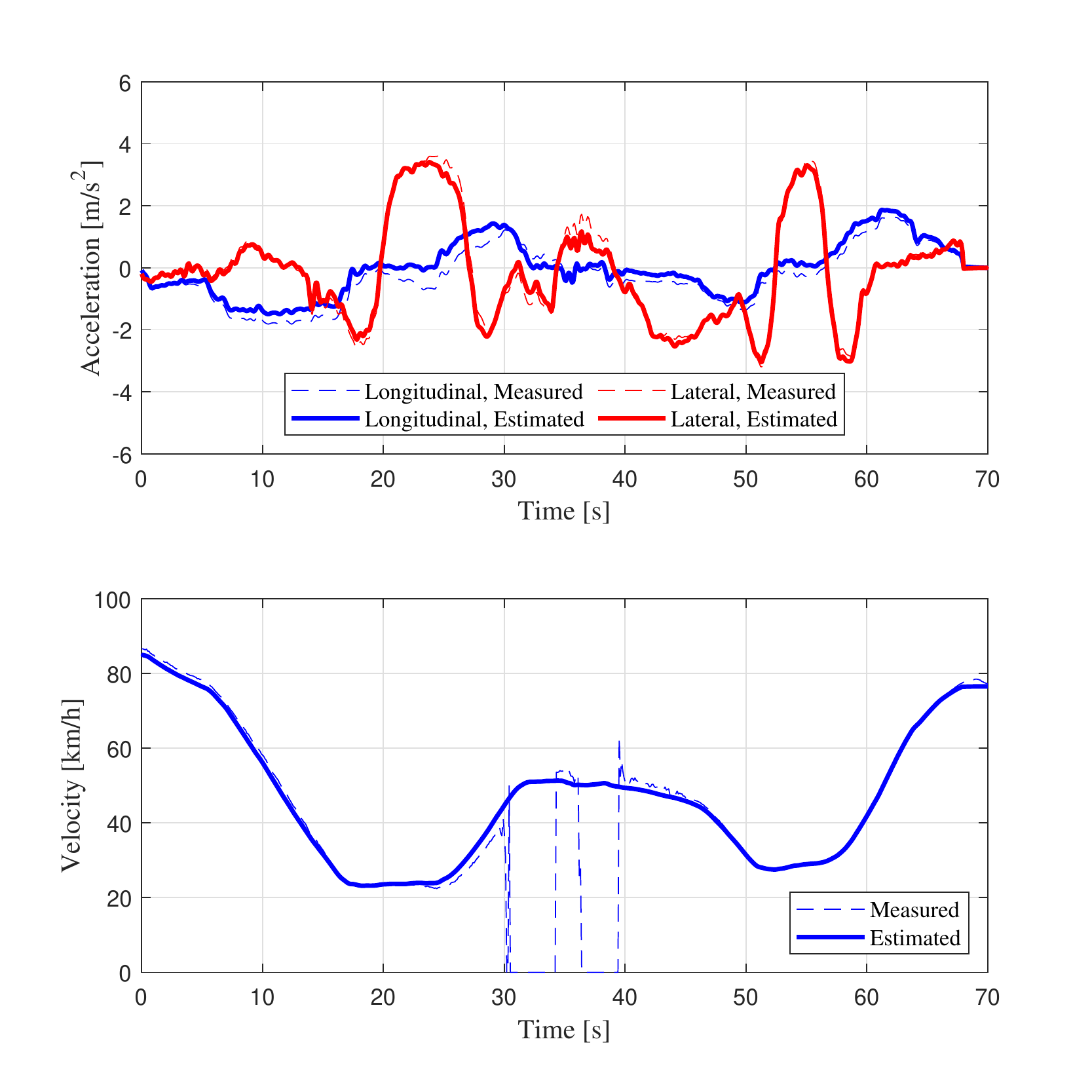}
    \caption{An example of the motion reconstruction method: estimated accelerations vs. IMU measurement (top) and estimated velocity vs. GPS measurement (bottom).}
    \label{fig:av_est}
\end{figure}

The following example is used to demonstrate the effectiveness of our proposed method for reconstructing vehicle motion during the test runs. The measured and estimated trajectories are compared in Fig. \ref{fig:traj_est} and the accelerations and velocities in Fig. \ref{fig:av_est}. Using our optimization-based method, the estimated trajectory exhibits a stable curvature when the measurement is of poor quality or unavailable. The estimated accelerations follow the measurements closely in general. The mismatch in longitudinal acceleration is more significant, possibly because of the elevation changes along the test route. The exit ramp where the initial braking phase happened shows a downward slope so that gravity is projected along the vehicle's longitudinal axis. Combining with positional measurement helped mitigate such effects. \par
A total of 31 test runs have been recorded during the experiment while in one other case, the participant did not follow the correct test route. Among these, 14 are considered usable. The decision is primarily based on the fact that the minimum speed of the vehicle is over \SI{18}{\kilo\meter\per\hour}. It suggests that the test vehicle was not required to yield when entering either of the two roundabouts. The total acceleration energy calculated with the reconstructed motion profiles is on average 12.2\% lower than that from the IMU signals. This amount of reduction accounts for factors including road elevation change, rotation of the vehicle body (roll and pitch), and noise. Without knowing the ground truth, we cannot rule out the possibility that the resulting motion profiles contain underestimated accelerations at some parts. However, we accept this risk because it then indicates a higher performance level of human drivers. By comparing to such a baseline performance, we avoid claiming an exaggerated advantage of automated vehicles. \par


\subsection{Comparison with individuals}
\begin{figure*}
    \centering
    \includegraphics[trim={80pt 30pt 60pt 20pt}, clip, width=\linewidth]{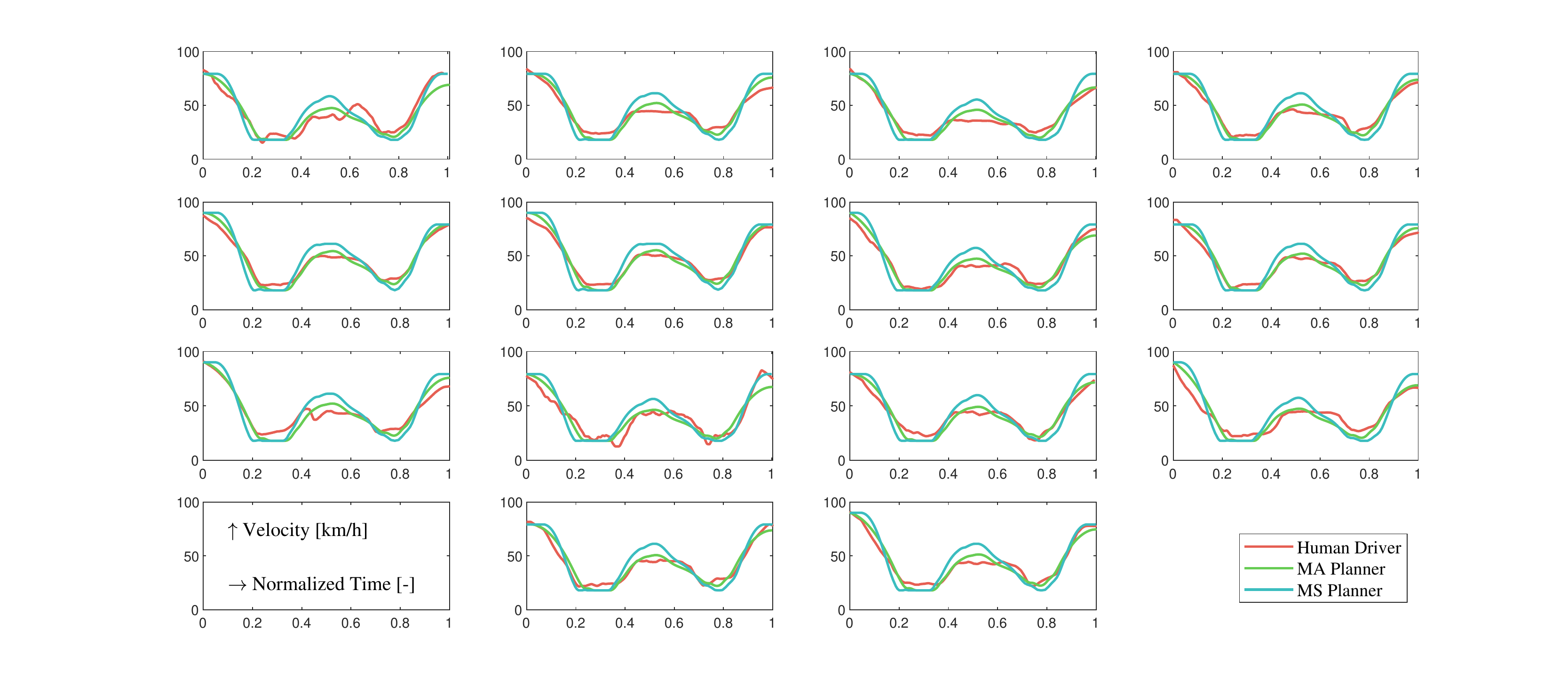}
    \caption{A collection of velocity profiles in normalized time, where the human driving data is compared with the optimized motion plans that have the same duration.}
    \label{fig:compare_vel}
\end{figure*}
For each valid test run, we generated optimization-based motion profiles using the motion planning algorithm described in Section \ref{sec:planner} with the two different objective functions. The weight on duration has been adjusted accordingly in order to obtain an approximate matching of travel time between the human driver and the motion planner. Because minimal instructions were given to the participants regarding how they should drive, the initial velocities at the start of the run are different. Hence we also adjusted the initial velocity in the motion planner to ensure a fair comparison. \par
\begin{figure*}
    \centering
    \includegraphics[trim={80pt 30pt 60pt 30pt}, clip, width=\linewidth]{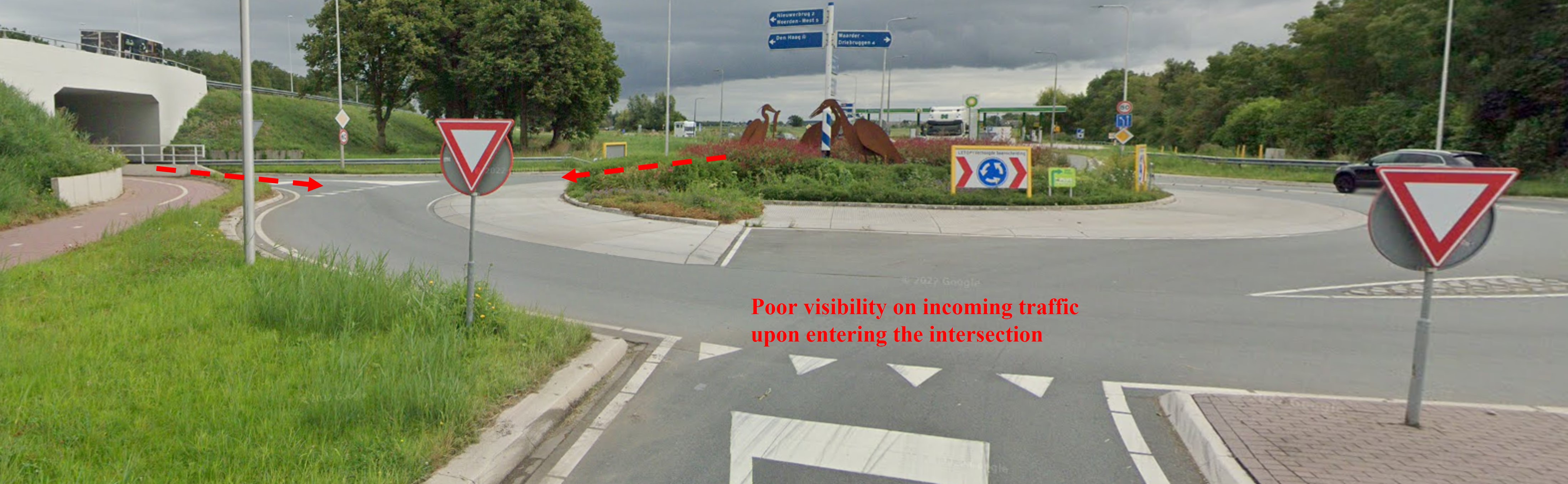}
    \caption{The visibility condition when entering the first roundabout on the test route. Vehicles entering from two other directions, represented by the arrows, only become visible at a distance of approximately \SI{30}{\meter}. Image source: Google Street View}
    \label{fig:vis_entry}
\end{figure*}

The velocity profiles of human drivers and motion planners are provided in Fig. \ref{fig:compare_vel}. Several differences between human drivers and our motion planners can be observed in multiple examples. First, most human-driven runs exhibit a more significant speed reduction at the beginning of the run. This is because of the drivers' attempt to save control effort by letting the vehicle coast for a reasonable distance ahead of an intersection as long as it does not impede other road users. Another potential cause of this difference is that the drivers were maintaining a wider safety margin when approaching an intersection with a higher initial speed. At the end of this braking period, the vehicle would enter RB1 where it needs to give way to vehicles traveling inside. By reducing speed in advance, the drivers have more time to observe the traffic situation and react safely. This is especially relevant because of the poor visibility at this roundabout as depicted by Fig. \ref{fig:vis_entry}). It is hardly possible for the drivers following our test route to detect other road users entering the roundabout from behind occlusions. Therefore, the participants could have decided to decelerate earlier to ensure comfort for the unforeseen case where they need to come to a full stop at the entry to RB1. Inside RB1, however, most drivers adopted a higher speed than the motion planners. The drivers might have overestimated the time loss or they were concerned about the following vehicle. Alternatively, they subconsciously tried to minimize fuel consumption and mechanical wear in the braking system. Again, the driving effort could have also played a role here because passing the roundabout with a lower speed means more pedal inputs. \par
In the RLR sector, the human drivers reached a lower top speed that is well below the imposed limit of 60 km/h. We believe this difference is primarily caused by the damaged pavement at the left-hand turn. The participants drove here at a lower speed in order to maintain good vertical comfort while the motion planners are developed under the assumption of ideal pavement conditions. Limited by this, the drivers would not attempt to reach the speed limit before the next right-hand turn that lies less than \SI{100}{\meter} ahead, where the vehicle needs to slow down again. \par
Finally, upon entering RB2, the participants often chose to decelerate later and more aggressively. In contrast to the first one, the second roundabout is located in an open area where oncoming traffic can be observed with little obstruction. This supports the discussion above that the visibility condition at an intersection may influence human drivers' choices of speed. Meanwhile, the feature of a higher speed by human drivers inside RB2 is similar to what was observed at RB1. After leaving RB2, most participants commanded a less aggressive acceleration than the motion planners. This could be relevant to the arguments above that human drivers may consider fuel consumption in their decision-making. \par

\subsection{Group performance}

\begin{figure}
    \centering
    \includegraphics[trim={20pt 00pt 30pt 10pt}, clip, width=\linewidth]{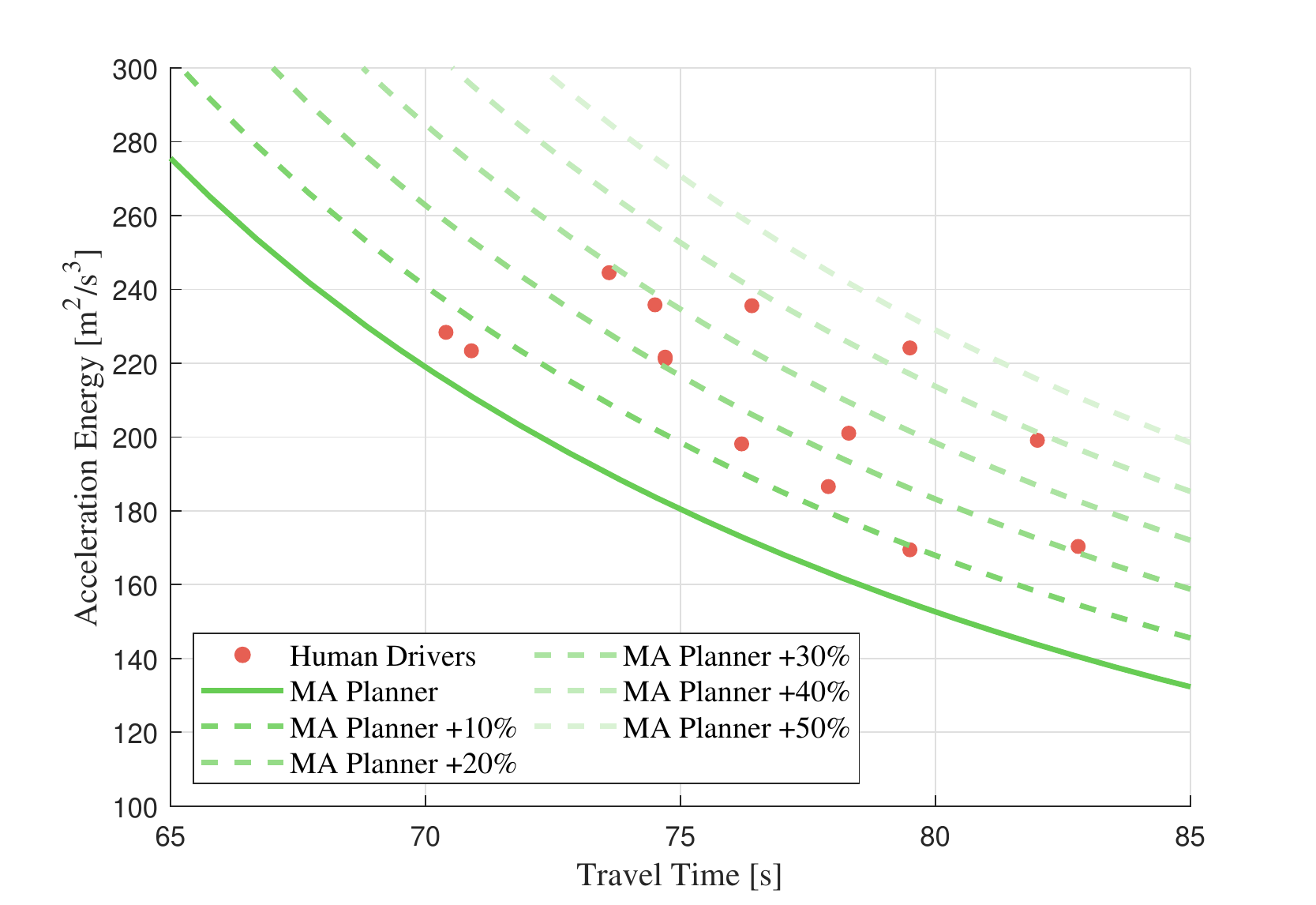}
    \caption{The performance of human drivers in improving acceleration comfort and minimizing travel time. The motion planner optimizing these two factors is visualized for comparison purposes. The dash lines are the contours corresponding to amplified discomfort when consuming the same amount of time.}
    \label{fig:compare_ac}
\end{figure}

  \begin{figure}
    \centering
    \includegraphics[trim={20pt 00pt 30pt 10pt}, clip, width=\linewidth]{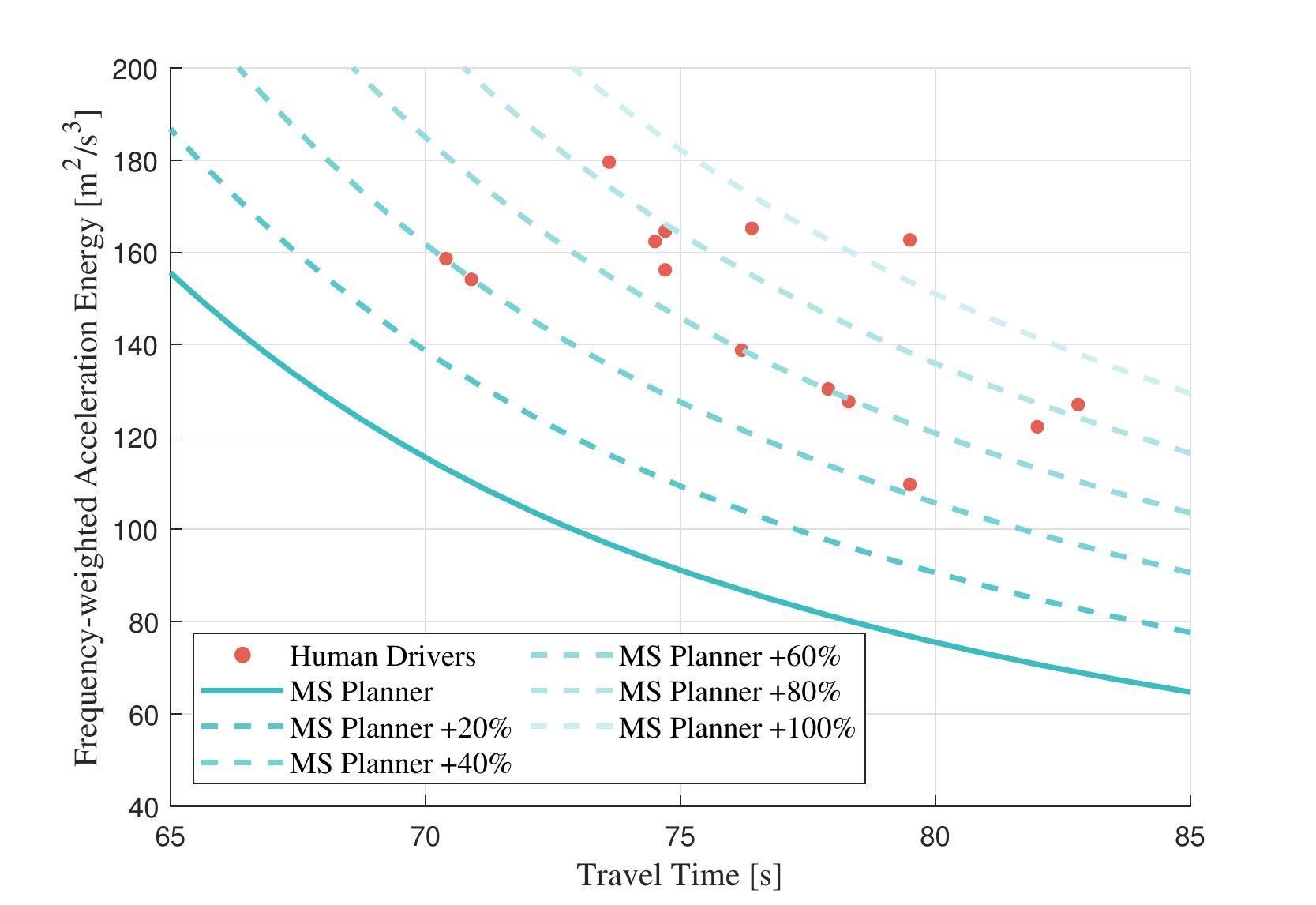}
    \caption{The performance of human drivers in mitigating motion sickness and minimizing travel time. The motion planner optimizing these two factors is visualized for comparison purposes. The dash lines are the contours corresponding to amplified predicted sickness levels when consuming the same amount of time.}
    \label{fig:compare_ms}
\end{figure}

\begin{figure}
    \centering
    \includegraphics[trim={30pt 00pt 30pt 10pt}, clip, width=\linewidth]{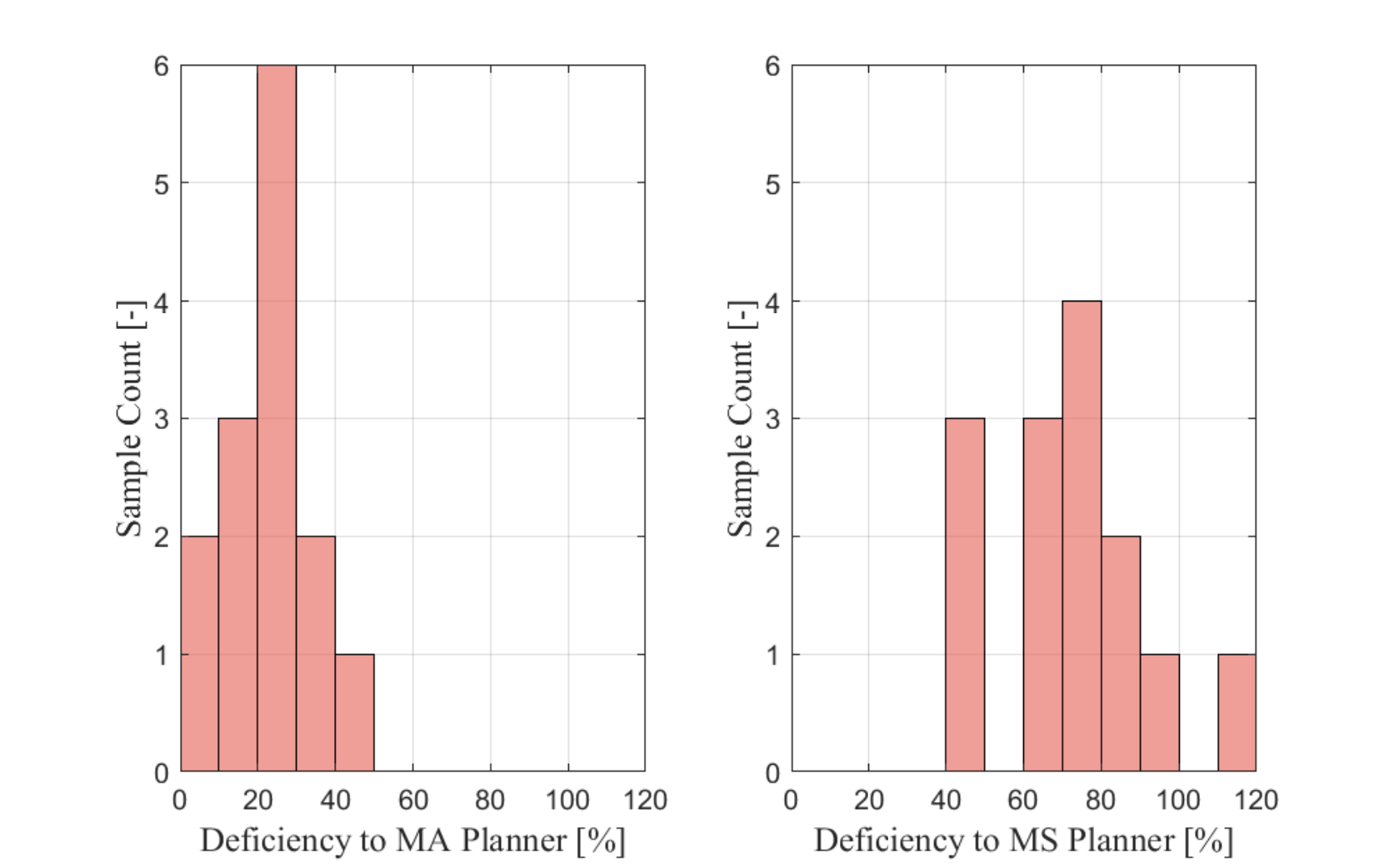}
    \caption{Distribution of human drivers' performance deficiency to the corresponding optimization-based motion planners.}
    \label{fig:histo_hd}
\end{figure}

The performance of human drivers is evaluated with the objectives of the motion planners: travel time and acceleration energy with or without frequency weighting. On average, the participants took \SI{76.5}{\second} to complete the test run while the average acceleration energy is \SI{211.4}{\square\meter\per\cubic\second} and the average frequency-weighted acceleration energy is \SI{147.1}{\square\meter\per\cubic\second}. In Fig. \ref{fig:compare_ac}, the human driving performance is compared with the MA planner with the acceleration energy being the comfort criterion. The participants exhibited up to 50\% more discomfort than the MA planner when consuming the same amount of time. The average disadvantage is 23.5\% with the majority (11 out of 14) lying between 0 and 30\%. It suggests that human drivers are reasonably good at planning vehicle motions that reduce the accelerations when they are experienced with the route. In contrast, human drivers' performance in mitigating motion sickness is significantly worse as suggested by Fig. \ref{fig:compare_ms}. The best-performing human driver still inflicted 20\% higher frequency-weighted acceleration energy than the MS planner while the average deficiency is 70.2\%. We highlight the distribution of performance deficiency with Fig. \ref{fig:histo_hd}. The more significant difference in mitigating motion sickness is probably caused by the human driver's lower susceptibility to motion sickness. When they cannot sense certain driving behaviors that give rise to motion sickness among the passengers, it is unlikely that they are able to avoid such behaviors. \par

\subsection{Frequency-domain Comparison}
\begin{figure*}
    \centering
    \includegraphics[trim={80pt 30pt 60pt 25pt}, clip, width=\linewidth]{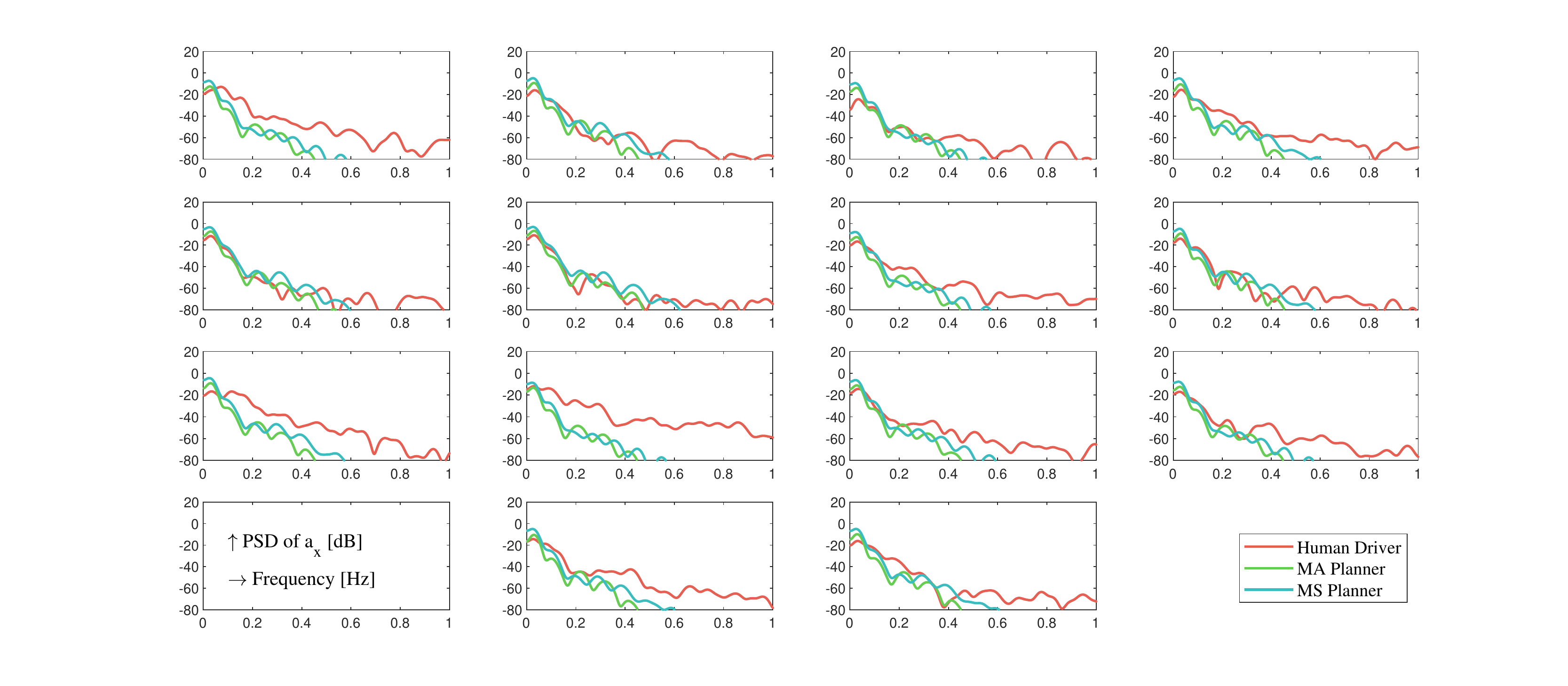}
    \caption{A collection of power spectral density in longitudinal accelerations, where the human driving data is compared with the optimized motion plans that have the same duration.}
    \label{fig:psd_ax}
\end{figure*}
\begin{figure*}
    \centering
    \includegraphics[trim={80pt 30pt 60pt 25pt}, clip, width=\linewidth]{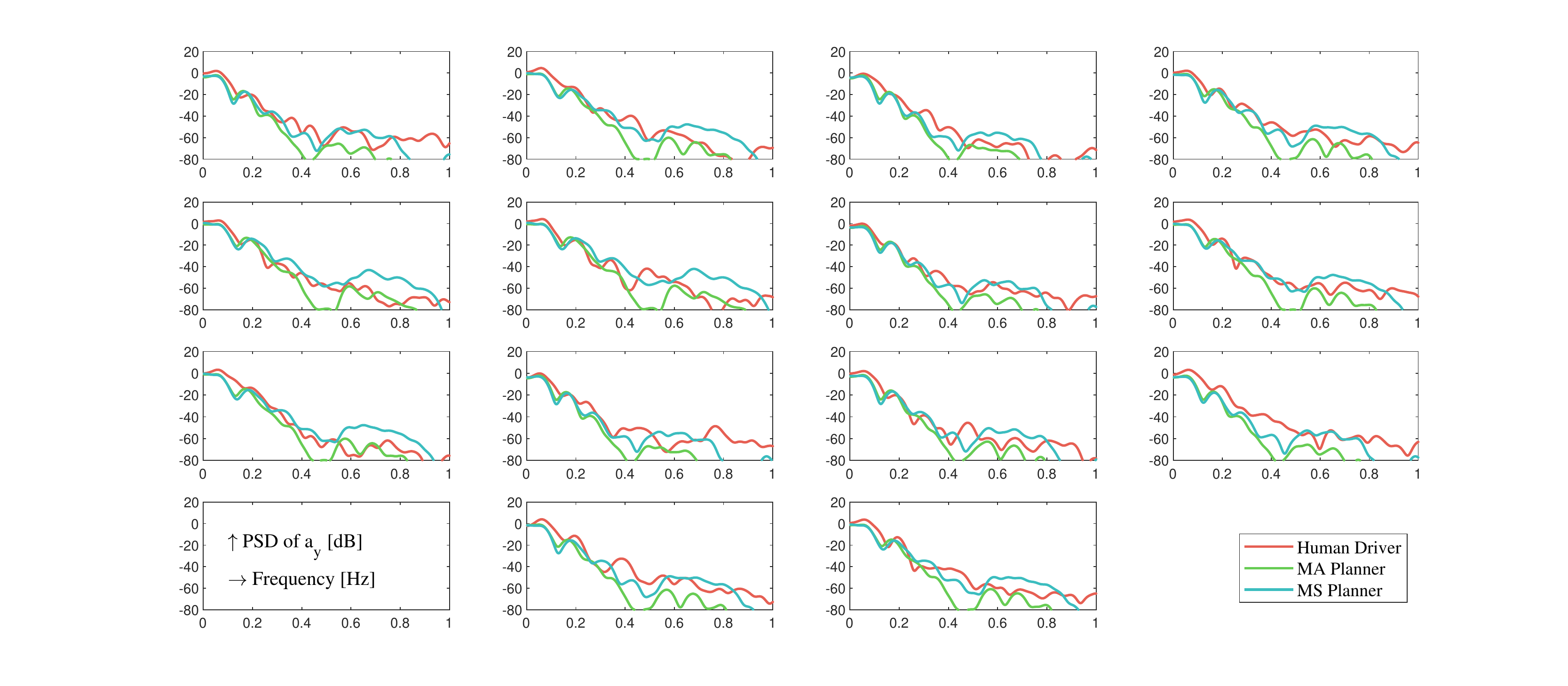}
    \caption{A collection of power spectral density in lateral accelerations, where the human driving data is compared with the optimized motion plans that have the same duration.}
    \label{fig:psd_ay}
\end{figure*}

The distribution of acceleration energy in the frequency domain is closely related to the incidence of motion sickness. The frequency components in the acceleration data from human drivers are presented in Fig. \ref{fig:psd_ax}, \ref{fig:psd_ay} in comparison with the motion plans from our optimization-based motion planners. Human drivers have more longitudinal acceleration energy distributed in the range above \SI{0.2}{\hertz}, which reflects more fluctuations in speed. These fluctuations could be perceived as uncomfortable or disturbing rather than nauseogenic because they are outside the most sensitive frequency range. We suspect that smooth control of vehicle speed is challenging for human drivers. Intuitively, it is physically demanding for human drivers to ensure smooth pedal inputs. As discussed before, they may attempt to save input effort when it does not significantly hamper comfort. Moreover, smooth input does not necessarily mean smooth vehicle speed because of the characteristics of the powertrain and the braking system. For example, when the transmission controller selects a higher gear, more throttle input is required in order to maintain the previous acceleration level. The driver has to sense a drop in acceleration before starting to adjust his throttle input. It is possible that they would accept a drop in acceleration and keep the input unchanged. \par
In lateral acceleration, the difference between human drivers and motion planners appears to be less significant. A higher magnitude can be seen from most human drivers for the frequency range below 0.2 Hz. This is mainly due to their choice of a higher speed when driving inside RB1 and RB2 because the low-frequency range is more reflective of the sustained accelerations. Given the frequency sensitivity in motion sickness, this difference could contribute significantly to motion sickness among passengers. In the higher frequencies, however, human drivers did not exhibit significantly more energy. The characteristics of the steering system could be one of the reasons. In general, the lateral acceleration of the vehicle is approximately linear with respect to the steering wheel input for the most common range of dynamics in daily driving, given that the tires are in the linear range. The steering ratio further reduces the scaling ratio between the steering wheel input and the resulting lateral and yaw motion of the vehicle. The haptic feedback is another factor that facilitates the lateral control task. The drivers can feel the torque on the steering wheel, which is indicative of lateral acceleration. Although a similar relationship exists between the brake pedal force and longitudinal acceleration, human beings are in general more sensitive with their hands than feet. Hence is lateral acceleration more smoothly controlled than longitudinal acceleration. \par

\section{Conclusions}\label{sec:concl}
\subsection{Contributions}
This paper presents an experimental study to measure the performance of human drivers in balancing comfort and time efficiency. The driving data of publicly recruited participants have been collected while they were driving an instrumented vehicle through the predefined route. An optimization-based processing method has been used to obtain a more accurate estimate of the actual vehicle motion. The motion profiles are then compared to two optimization-based motion planners from our previous research that aim to improve motion comfort by minimizing acceleration discomfort or the predicted incidence of motion sickness. \par 
In the time domain, we observed certain differences between human drivers and motion planners and analyzed the potential reasons leading to such differences. The performance deficiency of human drivers may not be solely to blame on their inferior capability in planning and controlling vehicle motion. Rather, they may have included other factors such as input effort and fuel consumption in their decision-making. Nevertheless, their driving behavior could potentially lead to a significantly higher level of motion sickness than the corresponding motion planner. We assert that this phenomenon is related to their lower susceptibility to motion sickness, which means they are not sensitive to the features in vehicle motion that are perceived by passengers as nauseogenic. \par 
In terms of the average performance, the participants show approximately 23.5\% more acceleration energy or 70.2\% more frequency-weighted acceleration energy than the motion planners. These values are calculated with identical travel times. The findings in this study suggest that AVs could potentially overcome the challenges posed by motion sickness by significantly improving their level of motion comfort over average human drivers without sacrificing time efficiency. \par
Furthermore, we are preparing to publish the raw and processed data gathered through the experiment on an open-access data platform. The link to the dataset will be provided as soon as it is accepted to the platform. Sharing the research data from this study would benefit researchers in a relevant field and allow more potential value to be extracted. \par
\subsection{Limitations and Future Works}
Due to the limited resources, the number of participants is relatively small and the gender composition is male-dominant. Combined with the lack of control over the traffic situation, only 14 recorded runs were considered useful for our purpose. A larger participant group with more female participants would make the findings more convincing from a statistical point of view. In terms of test equipment, some participants appeared to have no experience with an automatic transmission. This partly offset the benefit of having one less pedal to learn. The hybrid powertrain could have further complicated the speed control task for the participants due to the switching of power sources between the combustion engine and the electric motor. We expect more naturalistic driving behavior if the participants were to drive their own vehicles. Nevertheless, this would cause more difficulty in finding participants due to the increased effort (e.g., mounting and testing of data acquisition system on multiple vehicles) and risk (e.g., damaging the participants' vehicle). Besides, the performance deficiency of human drivers might be overestimated. The elevation change and pavement conditions faced by the participants are not considered by the motion planners. Meanwhile, the accelerations might be underestimated when processing the measurement data. The actual extent of how these two factors influence the suggested performance difference could not be verified. On the test route, the use of a GPS sensor alone as a source of position information is insufficient due to the unstable satellite connection. In future research, we recommend measuring the vehicle's relative position within the lane as an additional source of positional information, which could improve the estimation quality of the vehicle trajectory. This could be done by, for example, mounting cameras on the sides of the test vehicle and calculating the distance with intrinsic camera parameters. Lastly, we recommend the experimental validation of the findings from this study in terms of subjective comfort evaluation. The demonstrated advantage of AVs in mitigating motion sickness is based on empirical prediction methods of motion sickness. It is highly helpful in promoting AVs if the benefit could be reproduced in a real-world setup with human subjects. \par


%





\ifCLASSOPTIONcaptionsoff
  \newpage
\fi



%
\bibliographystyle{IEEEtran}
\bibliography{references}

%

\begin{IEEEbiography}[{\includegraphics[width=1in,height=1.25in,clip,keepaspectratio]{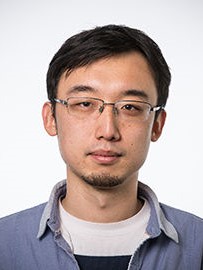}}]{Yanggu Zheng}
received the B.Sc. degree in Automotive Engineering from Tsinghua University, Beijing, China in 2015, and the M.Sc. degree (cum laude) from Delft University of Technology, Delft, the Netherlands in 2018. He is currently pursuing the Ph.D. in the Section of Intelligent Vehicles, Department of Cognitive Robotics, at Delft University of Technology. His research focuses on the application of optimization-based control and motion planning methods on automated vehicles for safety and comfort.
\end{IEEEbiography}

\begin{IEEEbiography}[{\includegraphics[width=1in,height=1.25in,clip,keepaspectratio]{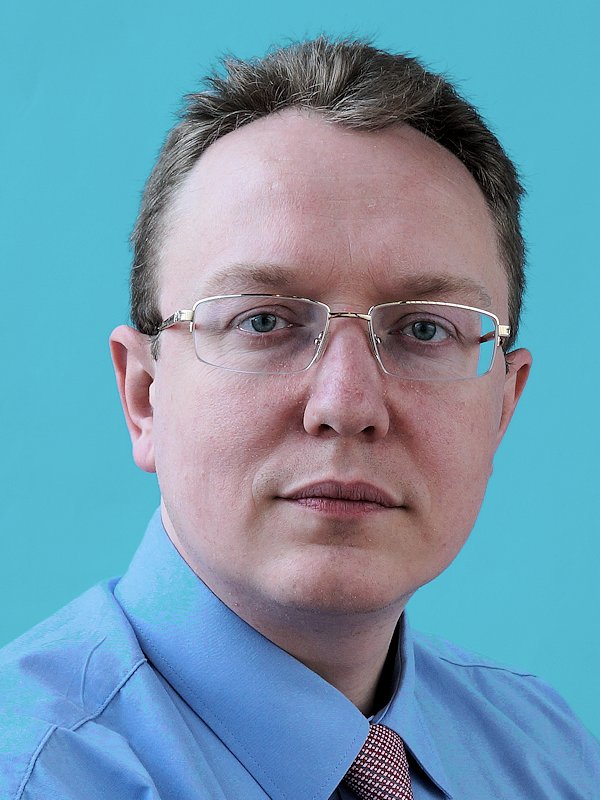}}]{Barys Shyrokau}
received the DiplEng degree (cum laude), 2004, in Mechanical Engineering from the Belarusian National Technical University, and the joint Ph.D. degree, 2015, in Control Engineering from Nanyang Technological University and Technical University Munich. He is an assistant professor in the Section of Intelligent Vehicles, Department of Cognitive Robotics, at Delft University of Technology and is involved in research related to vehicle dynamics and control, motion comfort, and driving simulator technology. He is a scholarship and award holder of FISITA, DAAD, SINGA, ISTVS, and CADLM.
\end{IEEEbiography}

\begin{IEEEbiography}[{\includegraphics[width=1in,height=1.25in,clip,keepaspectratio]{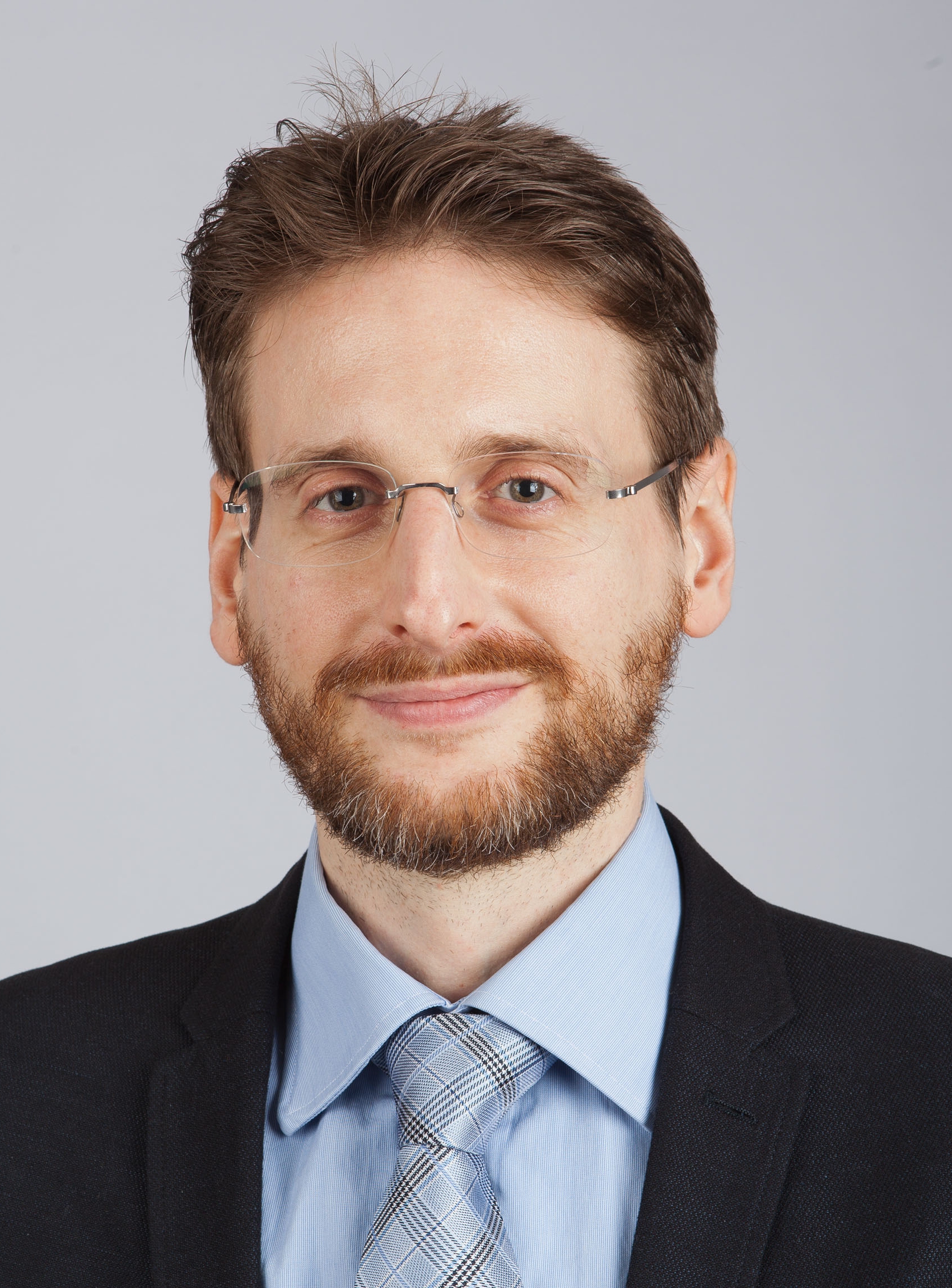}}]{Tamas Keviczky}
(Senior Member, IEEE) received the M.Sc. degree in electrical engineering from Budapest University of Technology and Economics, Budapest, Hungary, in 2001, and the Ph.D. degree from the Control Science and Dynamical Systems Center, University of Minnesota, Minneapolis, MN, USA, in 2005. He was a Post-Doctoral Scholar of control and dynamical systems with California Institute of Technology, Pasadena, CA, USA. He is currently a Professor with Delft Center for Systems and Control, Delft University of Technology, Delft, The Netherlands. His research interests include distributed optimization and optimal control, model predictive control, embedded optimization-based control and estimation of large-scale systems with applications in aerospace, automotive, mobile robotics, industrial processes, and infrastructure systems, such as water, heat, and power networks. He was a co-recipient of the AACC O. Hugo Schuck Best Paper Award for Practice in 2005. He has served as an Associate Editor for Automatica from 2011 to 2017 and for IEEE Transactions on Automatic Control since 2021.
\end{IEEEbiography}




\end{document}